\documentclass{article}

\usepackage{arxiv}
\usepackage[numbers]{natbib}

\usepackage[utf8]{inputenc} %
\usepackage[T1]{fontenc}    %
\usepackage{hyperref}       %
\usepackage{url}            %
\usepackage{booktabs}       %
\usepackage{amsfonts}       %
\usepackage{nicefrac}       %
\usepackage{microtype}      %
\usepackage{lipsum}		%
\usepackage{graphicx}
\usepackage{natbib}
\usepackage{doi}
\usepackage{caption}
\usepackage{subcaption}
\usepackage{xspace}
\usepackage{makecell}

\usepackage{amsmath}
\usepackage{xcolor}
\usepackage{listings}
\usepackage{makecell}
\usepackage{placeins}
\usepackage{xspace} %
\usepackage{morefloats} %
\usepackage{algorithmic}
\usepackage[lined,boxed,ruled,commentsnumbered]{algorithm2e}
\usepackage{placeins}
\usepackage{listings}

\newcommand{\ceinv}{CE$^{-1}$}

\newcommand\certain{\texttt{certain}\xspace}
\newcommand\fuzzy{\texttt{fuzzy}\xspace}

\title{Fuzzy Overclustering: Semi-Supervised Classification of Fuzzy Labels with Overclustering and Inverse Cross-Entropy}

\date{} 					%

\author{ \href{https://orcid.org/0000-0002-6945-5957}{\includegraphics[scale=0.06]{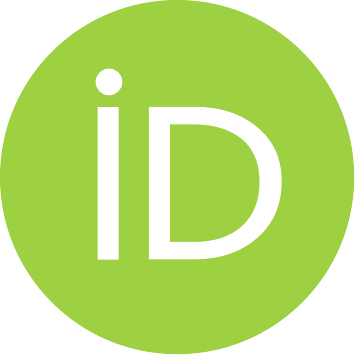}\hspace{1mm}Lars Schmarje}\thanks{Corresponding Author} \\
	Multimedia Information Processing Group\\
	Kiel University\\
	24118 Kiel, Germany \\
	\texttt{las@informatik.uni-kiel.de} \\
		\And
	\href{https://orcid.org/0000-0002-5118-145X}{\includegraphics[scale=0.06]{orcid.pdf}\hspace{1mm}Johannes Brünger} \\
	Multimedia Information Processing Group\\
	Kiel University\\
	24118 Kiel, Germany \\
	\texttt{jbr@informatik.uni-kiel.de} \\
		\And
	\href{https://orcid.org/0000-0002-4159-1367}{\includegraphics[scale=0.06]{orcid.pdf}\hspace{1mm}Monty Santarossa} \\
	Multimedia Information Processing Group\\
	Kiel University\\
	24118 Kiel, Germany \\
	\texttt{msa@informatik.uni-kiel.de} \\
		\And
	\href{https://orcid.org/0000-0002-6603-9907}{\includegraphics[scale=0.06]{orcid.pdf}\hspace{1mm}Simon-Martin Schröder} \\
	Multimedia Information Processing Group\\
	Kiel University\\
	24118 Kiel, Germany \\
	\texttt{sms@informatik.uni-kiel.de} \\
			\And
	\href{https://orcid.org/0000-0002-7851-9107}{\includegraphics[scale=0.06]{orcid.pdf}\hspace{1mm}Rainer Kiko } \\
	 Laboratoire d'Océanographie de Villefranche\\
	 Sorbonne Université \\
	 06230 Villefranche-sur-Mer, France;\\
	\texttt{rainer.kiko@obs-vlfr.fr} \\
	\And
	\href{https://orcid.org/0000-0003-4398-1569}{\includegraphics[scale=0.06]{orcid.pdf}\hspace{1mm}Reinhard Koch} \\
	Multimedia Information Processing Group\\
	Kiel University\\
	24118 Kiel, Germany \\
	\texttt{rk@informatik.uni-kiel.de} \\
}

\renewcommand{\headeright}{}
\renewcommand{\undertitle}{}
\renewcommand{\shorttitle}{Fuzzy Overclustering}

\hypersetup{
pdftitle={Fuzzy Overclustering: Semi-Supervised Classification of Fuzzy Labels with Overclustering and Inverse Cross-Entropy},
pdfauthor={Lars Schmarje},
pdfkeywords={Semi-supervised, ambiguous labels},
}

\begin{document}
\maketitle
	
	\begin{abstract}
Deep learning has been successfully applied to many classification problems including underwater challenges.
	However, a long-standing issue with deep learning is the need for large and consistently labeled datasets.
	Although current approaches in semi-supervised learning can decrease the required amount of annotated data by a factor of 10 or even more, this line of research still uses distinct classes.
	For underwater classification, and uncurated real-world datasets in general, clean class boundaries can often not be given due to a limited information content in the images and transitional stages of the depicted objects.
	This leads to different experts having different opinions and thus producing fuzzy labels which could also be considered ambiguous or divergent.
	We propose a novel framework for handling semi-supervised classifications of such fuzzy labels.
    It is based on the idea of overclustering to detect substructures in these fuzzy labels.
	We propose a novel loss to improve the overclustering capability of our framework and show the benefit of overclustering for fuzzy labels.
    We show that our framework is superior to previous state-of-the-art semi-supervised methods when applied to real-world plankton data with fuzzy labels.
	Moreover, we acquire 5 to 10\% more consistent predictions of substructures. 
	\footnote{The source code is avaialbe at \url{https://github.com/Emprime/FuzzyOverclustering}. The datasets are avaiable at \url{https://doi.org/10.5281/zenodo.5550918}.}
\end{abstract}
	
	\section{Introduction}

    Over the past years, we have seen the successful application of deep learning to many underwater computer vision problems~\cite{deep_fish,underwater_coral,underwater_cabel,temerateFish}.
    Automatic analysis of underwater data allows us to monitor ecological changes by evaluating large amounts of for example plankton data~\cite{consistent_planktonic,sinking_particles}.
    While it is relatively easy to create a lot of underwater image data, its analysis is time-consuming and thus expensive because the annotation requires trained taxonomists.
	The possible reasons for this issue include the huge amounts of data, the high imbalance between classes and the variability of annotations~\cite{merIssues}.

\begin{figure}
	\centering
\includegraphics[width=0.43\textwidth]{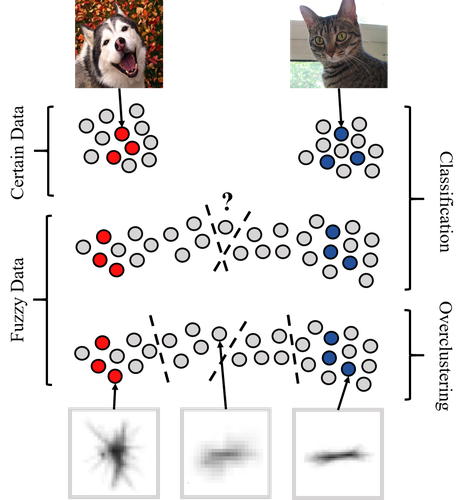}
\caption{Illustration of \fuzzy data and overclustering---The grey dots represent unlabeled data and the colored dots labeled data from different classes.
		The dashed lines represent decision boundaries.
		For \certain data, a clear separation of the different classes with one decision boundary is possible and both classes contain the same amount of data (\textbf{top}).
		For \fuzzy data determining a decision boundary is difficult because of intermediate datapoints between the classes (\textbf{middle}).
		These \fuzzy datapoints  can often not be easily sorted into one consistent class between annotators.
		If you overcluster the data, you get smaller but more consistent substructures in the \fuzzy data (\textbf{bottom}).
		The images illustrate possible examples for \certain data (cat \& dog) and \fuzzy plankton data (trichodesmium puff and tuft).
		The center plankton image was considered to be trichodesmium puff or tuft by around half of the annotators each.
		The left and right plankton image were consistently annotated.}\label{fig:ideav2}
\end{figure}

	In underwater classification, domain experts often differ in their annotations~\cite{merIssues,benthic_uncertainty,benthicIsis}.
	This issue arises due to the following reasons:
	Firstly, automatically captured underwater images often have a lower quality than images taken manually by humans.
	This difference in quality arises for example due to the underwater lighting conditions and no manual corrections to e.g. insufficient sharpness or not centering the target inside the focus.
	For example the analyis of benthic images can suffer from these issues~\cite{benthic_uncertainty,benthicIsis}.
	Even in the best scenario, a single image generally does not contain most of the information needed for a clear identification (e.g., three-dimensional configuration, minute morphological details, fluorescence).
	Secondly, intermediate stages actually exist between classes~\cite{morphocluster}. 
	For example, in Figure \ref{fig:ideav2} we show two different physical appearances (puff \& tuft) of trichodesmium, while the dataset also contains intermediate stages between these two~classes.
	This issue of different annotations is also known as \emph{intra-} and \emph{inter-observer} variability~\cite{noisy-labels-comparison} and is common in many biological and medical application fields~\cite{tailception,benthicIsis,schmarje2019,eye-fuzzy,cancergrading,medicinecrowdsource,planktonUncertain,benthic_uncertainty}.
	Even in a curated dataset~\cite{deep_fish}, we quote Tarling et al. who state ''there will very likely be inaccuracies, bias, and even inconsistencies in the labeling which will have affected the training capacity of the model and lead to discrepancies between predictions and ground truths''~\cite{underwater_uncertainty}.
	When aggregating multiple annotations per image, we call the resulting label \fuzzy if we have different annotations between experts (non-zero variance), and \certain if all annotations agree with each other.
	The mathematical formulation of a \fuzzy label would be a unknown soft probability distribution $l$ for $k$ classes.
	The distribution $l \in (0,1)^k$ can only be approximated with a high cost e.g., by averaging over multiple annotations.
	
	Semi- and Self-Supervised Learning are promising approaches to decrease the needed amount of annotated data by a factor of 10 or even more~\cite{remixmatch,S4L,simclr}.
	These approaches leverage unlabeled data in addition to the normal labeled data to improve the training. 
	A common strategy is to define a pretext task like image rotation prediction~\cite{self-rotation} or mutual information maximization~\cite{iic} for pretraining.
	A broad overview of current trends, ideas and methods in semi-, self- and unsupervised learning is available in~\cite{survey}.
	However, this research mainly focuses on established curated classification datasets such as STL-10~\cite{stl-10}.
	In these datasets, a clear distinction between classes such as cats and dogs are given. 
	The hard partitioning of intermediate morphologies is not appropriate and does not allow the identification of substructures.
	We show that state-of-the-art semi-supervised algorithms are not well suited to handle \fuzzy labels.
	These algorithms expect only \certain labels as shown in the upper part of Figure \ref{fig:ideav2}.
	If we apply previous semi-supervised algorithms to \fuzzy data which include \fuzzy images, these algorithms arbitrarily assign undecidable images to one class (middle part of Figure \ref{fig:ideav2}).
	
	\textls[-13]{Noisy labels are a common data quality issue and are discussed in the literature~\mbox{\cite{noisy-labels-comparison,surveynoise,surveynoise2}}}.
	The fuzziness of labels is known as a special case of label noise that exist ``due to subjectiveness of the task for human experts or the lack of experience in annotator[s]''~\cite{surveynoise}.
	In contrast to us, most methods~\cite{self-noisy,temporal-ensembling,divide-mix} and literature surveys~\cite{noisy-labels-comparison,surveynoise,surveynoise2} interpret \fuzzy labels as corrupted labels.
	We argue that \fuzzy labels are valid signals derived from ambiguous images and that it is important to discover the substructures for real-world data handling~\cite{tailception,schmarje2019,eye-fuzzy,cancergrading,medicinecrowdsource,planktonUncertain}.
	
	Geng proposed to learn the label distribution to handle \fuzzy data~\cite{labeldistribution} and the idea was extended to the application of real-world images~\cite{deep-learn-label-distribution}.
	However, these methods are not semi-supervised and therefore depend on large labeled datasets.
	A variety of methods was proposed to handle \fuzzy data in a semi-supervised learning approach~\cite{fuzzy-semi-supervised,fuzzymeta,semi-supervised-soft}.
	These methods use lower-dimensional features spaces in contrast to images as input.
	Liu et al. proposed to use independent predictions of multiple networks as pseudo-labels for the estimation of the label distribution for photo shot-type classification~\cite{fuzzyPhotos}.
	We argue that the true label distribution is difficult to approximate and thus difficult to evaluate.
	We do not learn the label distribution but use clustering to identify substructures. 
	
	We propose \emph{Fuzzy Overclustering} (FOC) which separates the \fuzzy data into a larger number of visual homogeneous clusters (lower part, Figure \ref{fig:ideav2}) which can then be annotated very efficiently~\cite{morphocluster}.
	We will show on a Plankton dataset that state-of-the-art semi-supervised algorithms perform worse on \fuzzy data in comparison to our method FOC which explicitly considers \fuzzy images.
	Moreover, we will show that this leads to 5 to 10\% more self-consistent predictions of plankton data.
	
	One main idea is to rephrase the handling of \fuzzy labels as a semi-supervised learning problem by using a small set of \certain images and a large number of \fuzzy images that are treated as unlabeled data.
	This approach allows us to use the idea of overclustering from semi-supervised literature~\cite{iic,deep-cluster} and apply it to \fuzzy data.
	The difference to previous work is that we use overclustering not only to improve classification accuracy on the labeled data but improve the clustering and therefore the identification of substructures of \fuzzy data.
	We show that overclustering allows us to cluster the \fuzzy images in a more meaningful way by finding substructures and therefore allowing experts to analyze \fuzzy images more consistently in the future.
	
	We show the benefits of our method mainly on a plankton dataset which highlights the benefit for underwater classification.
	However, the issue of \fuzzy labels is neither limited to plankton data nor to underwater classification.
	On a synthetic dataset, we show a proof-of-concept for the generalizability of our model to other datasets.

	Our key contributions are:
	\begin{itemize}
		\item We identify an issue of semi-supervised algorithms that they do not work well with \fuzzy labels. 
		However, such \fuzzy labels occur regularly in underwater image classification e.g due to high natural variation of depicted objects which leads to a high inter- and intraobserver variability.
		\item We propose a novel framework for handling \fuzzy labels with a semi-supervised approach.
		This framework uses overclustering to find substructures in \fuzzy data and outperforms common state-of-the-art semi-supervised methods like FixMatch~\cite{fixmatch} on \fuzzy plankton data.
		\item We propose a novel loss, \emph{Inverse Cross-entropy} (\ceinv), which improves the overclustering quality in semi-supervised learning. 
		\item We achieve 5 to 10\% more self-consistent predictions on \fuzzy plankton data.
		
	\end{itemize}
	
	\section{Method}
	\label{sec:methods}
	Our framework Fuzzy Overclustering (FOC) aims at creating an overclustering for \fuzzy labels by using an auxiliary classification and not the other way round like previous literature~\cite{iic,deep-cluster}.
	In this section, we describe our framework in general and explain important parts in detail in the following subsections.	
	We use the following notation for the given semi-supervised classification task. 
	Our training data consists of the two subsets $X_l$ and $X_u$. $X_l$ is a labeled image dataset with images $x \in X_l$ and corresponding labels $y$. $X_u$ is an unlabeled image dataset, i.e., there is/exists no label for images $x \in X_u$.

	We generate three inputs  $x_1,x_2,x_3$ based on one image $x \in X_l \cup X_u$ depending on the availability of the corresponding label $y$.
	If $y$ is not available, the images $x_1$ and $x_2$ are augmented views of $x$ and $x_3$ is an augmented version of a random image $x' \in X_l \cup X_u$.
	If $y$ is available, $x_1$ is an augmented view of $x$, $x_2$ is a supervised augmentation (see Section \ref{subsec:aug}) and $x_3$ an inverse example.
	For the inverse example, we choose an image $x' \in X_l$ with a different label $y'$ ($y != y'$).
	We use an augmented version of this image as third input $x_3 = g_3(x')$ with augmentation $g_3$.
	We constraint the ratio from unlabeled to labeled data to a fixed ratio $r$ to improve the run time of the model (see Section \ref{subsec:restricted}).
	The inputs are processed by a neural network $\Phi$ which is composed of a backbone like ResNet50~\cite{resnet} and linear output prediction layers.
	Following~\cite{iic}, we call this linear predictors \emph{heads} and use them either as normal or overclustering heads.
	As output we use the soft-max classifications of these normal and overclustering heads.
	If $k_{GT}$ is the number of ground-truth classes a normal head outputs a probability for each of the  $k_{GT}$ classes.
	The overclustering head has $k$ output nodes with $k > k_{GT}$ and give probabilities for more clusters than ground-truth classes (overclustering).
	Both type of heads are therefore fully connected layers with softmax activation but of different output size.
	We can average the training over multiple independent heads per type as shown in~\cite{iic}.
	We use the notation $\Phi_{n_i}$ or $\Phi_{o_i}$ for the i-th normal or overclustering head respectively.
	An overview about the general pseudo code of FOC including the loss calculation is given in Algorithm \ref{alg:pseudo}.

\begin{figure}
	\centering
\includegraphics[width=0.6\textwidth]{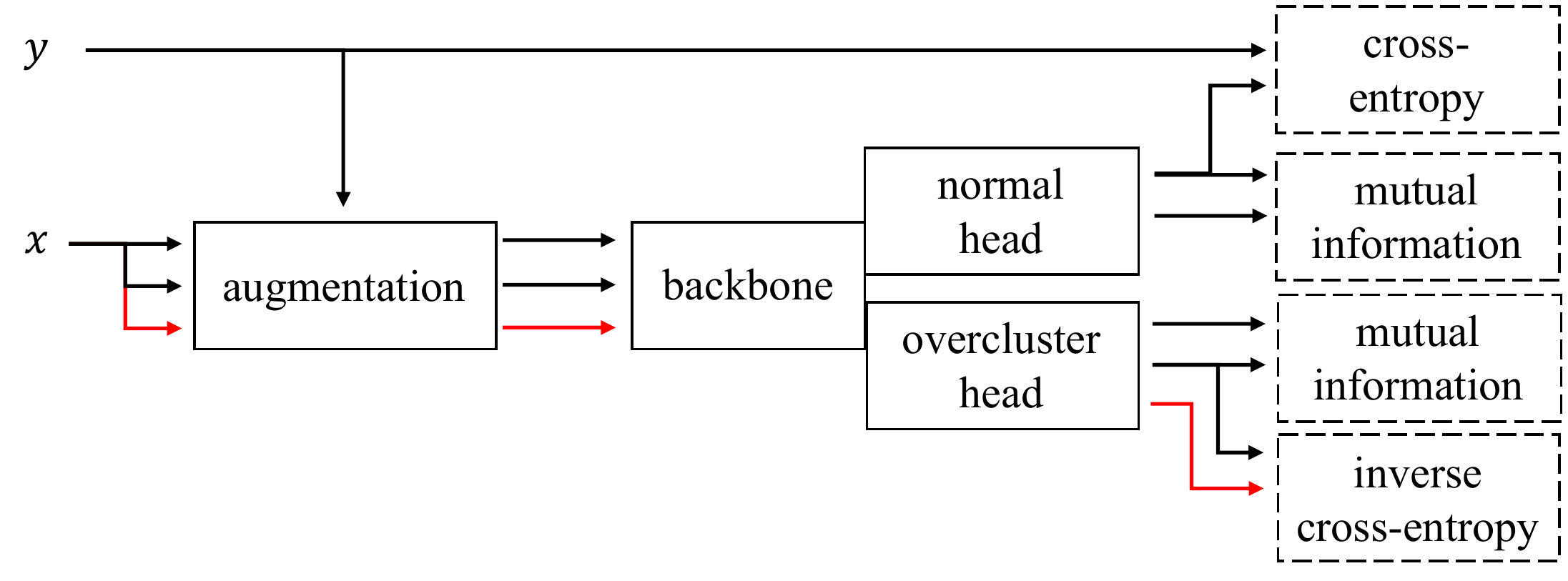}
\caption{Overview of our framework FOC for semi-supervised classification---The input image is $x$ and the corresponding label is $y$.
		The arrows indicate the usage of image or label information.
		Parallel arrows represent the independent copy of the information.
		The usage of the label for the augmentations is described in Section \ref{subsec:aug}.
		The red arrow stands for an inverse example image $x'$ with a different label than $y$.
		The output of the normal and the overclustering head have different dimensionalities.
		The normal head has as many outputs as ground-truth classes exist ($k_{GT}$) while the overclustering head has $k$ outputs with  $k > k_{GT}$.
		The dashed boxes on the right side show the used loss functions.
		More information about the losses inverse cross-entropy and mutual information can be found in Sections \ref{subsec:ceinv} and \ref{subsec:mi} respectively.}\label{fig:pipeline-1}
\end{figure}

	For both heads the loss is different but can be written as the weighted sum of an unsupervised and a supervised loss as follows:
	\begin{equation}
	\label{eq:loss}
	\mathcal{L} = \lambda_s \mathcal{L}_s + \lambda_u \mathcal{L}_u
	\end{equation}
	$\mathcal{L}_s$ is cross-entropy ($\mathcal{L}_{CE}$) for the normal head  and our novel \ceinv~loss ($\mathcal{L}_{CE^{-1}}$) for the overclustering head (see Section \ref{subsec:ceinv}).	
	For both heads $\mathcal{L}_u$ is the mutual information loss $\mathcal{L}_{MI}$ (see Section \ref{subsec:mi}). 
	An illustration of the complete pipeline is given in Figure \ref{fig:pipeline-1}.
	We initialize our backbones with pretrained weights and can therefore directly use RGB images as input. 
	For further implementation details see Section \ref{subsec:implementation}.

	If we use FOC with $\lambda_s = 0$ and without supervised augmentations our model is comparable to the pretext task of Invariant Information Clustering (IIC)~\cite{iic}.
	We can use this configuration as a warm-up to pretrain the weights.
	During the evaluation, we will refer to using the pretext task for IIC and the warm-up of FOC synonymously.
	Our framework FOC can also be used to perform standard unsupervised clustering.
	The details about unsupervised clustering and a comparison to previous literature is given in the supplementary. 
	
	\begin{algorithm}
     \KwData{Batch of images of size $b$ from labeled image data $X_l$ and unlabeled image data $X_u$ }
     \KwResult{calculate loss value for one given batch for a network $\Phi$ with $n$ normal and overclustering heads }
     $L$: matrix of size $b$ $\times$ $2n$\;
     \tcc{iterate over batch}
     \For{$i\leftarrow 0$ \KwTo $b$}{
     $x \leftarrow$ \emph{i}-th image in batch\;
      \eIf{label $y$ for image $x_i$ available}{
       $x_1 \leftarrow g_1(x)$ with random augmentation $g_1$\;
       \tcc{Supervised augmentation defined in Section \ref{subsec:aug}}
        $x_2 \leftarrow g_2(x)$ with supervised augmentation $g_2$\;
        \tcc{Inverse example defined in Section \ref{sec:methods}}
       $x_3 \leftarrow g_3(x')$ with random augmentation $g_3$ and inverse example $x'$\;

       }{
       $x_1 \leftarrow g_1(x)$ with random augmentation $g_1$\;
        $x_2 \leftarrow g_2(x)$ with random augmentation $g_2$\;
       $x_3 \leftarrow g_3(x')$ with random augmentation $g_3$ and random image \emph{x}'\;
      }
      \tcc{iterate over heads}
      \For{$j\leftarrow 0$ \KwTo $n$}{
        calculate forward pass for outputs $\Phi_{n_j}$ and $\Phi_{o_j}$\;
        \tcc{CE loss for normal head}
        L[i,j] $\leftarrow  \mathcal{L}_{CE}(\Phi_{n_j}(x_i), l_i)$ with $l_i$\;
        \tcc{\ceinv~loss for overclustering head}
        L[i,j+n] $\leftarrow \mathcal{L}_{CE^{-1}}(x_1,x_2,x_3)$ with 	Equation \eqref{eq:ceinv} \;

      }
     }
     \tcc{calculate loss}
     $\mathcal{L}_s$ $\leftarrow$ average supervised loss across heads and batch from $L$\;
     $\mathcal{L}_u$  $\leftarrow$ unsupervised MI loss across batch with Equations \eqref{eq:mi1} and \eqref{eq:mi2}\;
     $\mathcal{L} \leftarrow  \lambda_s \mathcal{L}_s +  \lambda_u \mathcal{L}_u$\;
     \caption{Pseudocode for our method Fuzzy Overclustering}     
     \label{alg:pseudo}
    \end{algorithm}
    
    \vspace{-12pt}
	\subsection{Inverse Cross-Entropy (\ceinv)}
	\label{subsec:ceinv}

	Inverse Cross-Entropy is a novel supervised loss for an overclustering head and one of the key contributions of this work.
	The loss is needed to use the label information for an overclustering head.	
	For normal heads, we can use cross-entropy (CE) to penalize the divergence between our prediction and the label.
	We can not use CE directly for the overclustering heads since we have more clusters than labels and no predefined mapping between the two.
	However, we know that the inputs $x_1 / x_2$ and $x_3$ should not belong to the same cluster. 
	Therefore, our goal with \ceinv~is to define a loss that pushes their output distributions (e.g., $\Phi(x_1)$ and  $\Phi(x_3)$) apart from each other.
	
	Let us assume we could define a distribution that $\Phi(x_3)$ should not be.
	In short, an inverse distribution $\Phi(x_3)^{-1}$.
	If we had such a distribution we could use CE to penalize the divergence for example between $\Phi(x_1)$ and $\Phi(x_3)^{-1}$.

	One possible and easy solution for an inverse distribution is $\Phi(x_3)^{-1} = 1 - \Phi(x_3)$. 
	For a binary classification problem, $\Phi(x_3)^{-1}$ can even be interpreted as a probability distribution again. 
	This is not the case for a multi-class classification problem.
	We could use a function like softmax to cast $\Phi(x_3)^{-1}$ into a probability distribution but decided against it for three reasons.
	Firstly, we would penalize correct behavior. 
	For example in a three class problem with $\Phi_1(x_1) = 0.5 = \Phi_2(x_1)$ and $\Phi_3(x_3) = 1$ we only get $CE(\Phi(x_1), \Phi(x_3)^{-1}) = 0$ if $\Phi(x_3)^{-1}$ is not a probability distribution.
	Otherwise either $\Phi_1(x_3)^{-1}$ or $\Phi_2(x_3)^{-1}$ have to be real smaller than 1.
	Secondly, we are still minimizing the entropy of $\Phi(x_1)$ which leads to more confident predictions in semi-supervised learning~\cite{entropy-min,S4L,UDA,vat,mixmatch,remixmatch}.
	The proof is given in the supplementary. 
	Thirdly, it is easier and in practice, it is not needed. 
	For the input $i = (x_1, x_2, x_3)$, we define the cross-entropy inverse loss $\mathcal{L}_{CE^{-1}}$ as shown in Equation \eqref{eq:ceinv}.
	\begin{equation}
	\label{eq:ceinv}
	\begin{split}
	\mathcal{L}_{CE^{-1}}(i) & = 0.5 \cdot CE^{-1}(\Phi(x_1), \Phi(x_3)) \\
	& + 0.5 \cdot CE^{-1}(\Phi(x_2), \Phi(x_3)) \,\mbox{, with } \\
	CE^{-1}(p,q) & = - \sum_{c=1}^{k} p(c) \cdot ln(1-q(c))\,.
	\end{split}
	\end{equation}

	\subsection{Mutual Information (MI)}
	\label{subsec:mi}
	
	For the unlabeled data, we use the loss proposed by Ji et al. because it is calculated directly on the output clusters~\cite{iic}.
	Therefore similar images are pulled to the same clusters while \ceinv~pushes different images apart.
	For this purpose, we want to maximize the mutual information between two output predictions $\Phi(x_1), \Phi(x_2)$ with  $x_1, x_2$ images which should belong to the same cluster and $\Phi :  X \rightarrow [0,1]^k$ a neural network with $k$ output dimensions.
	We can interpret $\Phi(x)$ as the distribution of a discrete random variable $z$ given by $P(z = c | x) = \Phi_c(x)$ for $c \in \{1, \dots, k\}$ with $\Phi_c(x)$ the c-th output of the neural network.
	With $z,z'$ such random variables we need the joint probability distribution for $P_{cc'} = P(z=c, z'=c')$ for the calculation of the mutual information $I(z,z')$.
	\mbox{Ji et al.} propose to approximate the matrix $P$ with the entry $P_{cc'}$ at row $c$ and column $c'$ by averaging over the multiplied output distributions in a batch of size $n$~\cite{iic}.
	Symmetry of $P$ is enforced as shown in Equation \eqref{eq:mi1}.
	\begin{equation}
	\label{eq:mi1}
	P = \frac{Q + Q^T}{2} \mbox{ with } Q = \frac{1}{n} \sum_{i=1}^{n} \Phi(x_i) \cdot \Phi(x_i')^T     
	\end{equation}
	
	We can maximize our objective $I(z,z')$ with the marginals $P_c = P_{c'} = P(z = c)$ given as sums over the rows or columns as shown in Equation \eqref{eq:mi2}.
	\begin{equation}
	\label{eq:mi2}
	I(z,z') = \sum_{c=1}^{k}\sum_{c'=1}^{k} P_{cc'} \cdot ln \frac{P_{cc'}}{P_{c} \cdot P_{c'}}
	\end{equation}

	\subsection{Supervised Augmentations}
	\label{subsec:aug}
	
	In the unsupervised pretraining, we use the same image $x$ to create the two inputs $x_1 = g_1(x)$ and $x_2 = g_2(x)$ based on the augmentations $g_1$ and $g_2$. 
	Otherwise, without supervision, it is difficult to determine similar images. 
	However, if we have the label $y$ for $x$ we can use a secondary image $x' \in X_l$ with the same label to mock an ideal image transformation to which the network should be invariant.
	In this case we can create $x_2 = g_2(x')$ based on the different image.
	We call this \emph{supervised augmentation}.

	\subsection{Restricted Unsupervised Data}
	\label{subsec:restricted}
	
	Unlabeled data has a small impact on the results but drastically increases the runtime in most cases.
	The increased runtime is caused by the facts that we often have much more unlabeled data than labeled data and that a neural network runtime is normally linear in the number of samples it needs to process.
	However, unlabeled data is essential for our proposed framework and we can not just leave it out.
	We propose to restrict the unlabeled data to a fixed upper-bound ratio $r$ in every batch and therefore the unlabeled data per epoch.
	Detailed examples and experiments are given in the supplementary.	
	It is important to notice that we restrict only the unlabeled data per batch/epoch.
	While for one epoch the network will not process all unlabeled data, over time all unlabeled data will be seen by the network.
	We argue that the impact on training time negatively outweighs the small benefit gained from all unlabeled data per epoch.

	\section{Experiments}
	\label{sec:evaluation}
	
	We conducted our experiments mainly on a real-world plankton dataset.
	We used the common image classification dataset STL-10 as a comparison with only \certain labels and  a synthetic dataset for a proof-of-concept for the generalizability to other datasets.
	We compare ourselves to previous work and  make several ablations.
	Additional results like unsupervised clustering, more detailed ablations and further details are given in the supplementary material.

	\subsection{Datasets}
	\label{subsec:datasets}
	
	While the issue of \fuzzy labels is present in multiple datasets~\cite{tailception,schmarje2019,eye-fuzzy,cancergrading,medicinecrowdsource,planktonUncertain}, they are not well suited for evaluations.
	If we want to quantify the performance on \fuzzy labels, we need a dataset with very good \fuzzy ground-truth.
	This can only be achieved with a high cost e.g. by multiple annotations and thus is often not feasible.
	For all used datasets, we ensure that the labeled training data only consists of \certain images and that the \fuzzy images are used as unlabeled data.
	If we include \fuzzy labels in the labeled data which is used as guidance during training, this will lead to worse performance as illustrated in the ablations (Table \ref{tbl:comparisonSYN}).
	
	\subsubsection{Plankton}
	
	The plankton dataset contains diverse grey-level images of marine planktonic organisms.
	The images were captured with an Underwater Vision Profiler~5~\cite{uvp} and are hosted on EcoTaxa~\cite{ecotaxa}.
	In the citizen science project PlanktonID \footnote{ \url{https://planktonid.geomar.de/en}}, %
	each sample was classified multiple times by citizen scientists. 
	The data for the PlanktonID project is a subset of the data available on EcoTaxa~\cite{ecotaxa}.
	It was presorted to contain a more balanced representation of the available classes.
	The dataset consists of  12,280 images in originally 26 classes.
	We merged minor and similar classes so that we ended up with 10 classes.
	The class no-fit represents a mixture of left-over classes.
	The merging was necessary because some classes had too few images for current state-of-the-art semi-supervised approaches. 
	After this process, a class imbalance is still present with the smallest class containing about 4.16\% and the largest class 30.37\% of all samples.
	We use the mean over all annotations as the \fuzzy label.
	The citizen scientists agree on most images completely. 
	We call these images and their label \certain.
	However, about 30\% of the data has as least one disagreeing annotation.
	We call these images and their label \fuzzy and use the most likely class as ground-truth if we need a hard label for evaluation.
	The \fuzzy labeled images are used only as unlabeled data.
	More details about the mapping process, the number of used samples and graphical illustrations are given in the supplementary.
	
	\subsubsection{STL-10}
	
	STL-10 is a common semi-supervised image classification dataset~\cite{stl-10} and a subset of ImageNet~\cite{imagenet}.
	It consists of 5000 training samples and 8000 validation samples depicting everyday objects.
	Additionally, 100,000 unlabeled images are provided that may belong to the same or different classes than the training images.
	In contrast to the plankton and synthetic dataset, no labels are provided for the unlabeled data and no \fuzzy datapoints exist.
	We use this dataset only to illustrate the difference in the performance of FOC to previous semi-supervised methods.

	\subsubsection{Synthetic Circles and Ellipses (SYN-CE)}
	
	This dataset is a mixture of circles and ellipses (bubbles) on a black background with different colors.
	The 6 ground-truth classes are blue, red and green circles or ellipses.
	An image is defined as \certain if the hue of the color is 0 (red), 120 (green) or 240 (blue) and the main axis ratio of the bubble is 1 (circle) or 2 (ellipse).
	Every other datapoint is considered \fuzzy and the ground-truth label $l$ is calculated as the product of the interpolation of the color $p_c$ and the geometry $p_g$ distribution.
	More details are in the supplementary.
	The dataset consists of 1800 \certain and 1000 \fuzzy labeled images for train, validation and unlabeled data split.
	We look at three subsets: \emph{Ideal}, \emph{Real} and \emph{Fuzzy}.
	The \emph{Ideal} subset uses the maximal class of the \fuzzy label $l$ as a ground-truth class and represents the ideal case that we certainly know the most likely label to each image.
	For the \emph{Real} subset, the ground-truth classes in randomly picked with the distribution of the \fuzzy label $l$ and represent the real or common case. 
	For example due to only one annotation, the percentage that the label corresponds to the actual most likely class is linear to the \fuzzy label. 
	The \emph{Fuzzy} subset only uses \certain labeled images as training data and represent a cleaned training dataset.
	We will show that this handling of \fuzzy labels leads to a higher classification performance in comparison to the \emph{Real} dataset in \autoref{subsec:syn}.
	The \emph{Ideal} and the \emph{Real}  subset can be evaluated on the unlabeled data of the \emph{Fuzzy} subset with some overlap in the images. %
	
	\subsection{Implementation Details}
	\label{subsec:implementation}
	
	As a backbone for our framework, we used either a ResNet34 variant~\cite{iic} or a standard ResNet50v2~\cite{resnet}.
	The heads are single fully connected layers with a softmax activation function.
	Following~\cite{iic}, we use five randomly initialized copies for each type of head and repeat images per batch three times for more stable training.
	We alternated between training the different types of heads.
	The inputs are either sobel-filtered images or color images for pretrained networks.
	For the ResNet34 backbone, we use CIFAR20 (20 superclasses in CIFAR-100~\cite{cifar}) weights and for the ResNet50v2 backbone ImageNet~\cite{imagenet} weights. 
	We use in general $\lambda_s = 1 = \lambda_u$ and an unlabeled data restriction of $r = 0.5$.
	We call our Framework FOC-Light if we use $\lambda_u  = 0$ and no warm-up. 
	This means we do not use the loss introduced by~\cite{iic} and therefore also do not have to use their stabilization methods like repetitions.
	During the pretext task or warm-up and the main training, we train the framework with Adam and an initial learning rate of 1 $\times$ 10$^{-4}$ for 500 epochs.
	When switching from the pretext task to fine-tuning, we train only the heads for 100 epochs with a learning rate of 1 $\times$ 10$^{-3}$ before switching to the lower learning rate of 1 $\times$ 10$^{-4}$.
	The number of outputs for the overclustering head should be about 5 to 10 times the number of classes.
	The exact number is not crucial because it is only an upper bound for the framework. 
	We use 70 for STL-10 and 60 for the plankton dataset.
	We selected all hyperparameters heuristically based on the STL-10 dataset and did not change them for the plankton dataset.
	We used the recommended hyperparameters by the original authors for the previous methods.
	We compared with the following methods Semantic Clustering by Adopting Nearest neighbors (SCAN)~\cite{scan}, Information Invariant Clustering (IIC)~\cite{iic}, Mean-Teacher~\cite{mean-teacher}, Pi(-Model)~\cite{temporal-ensembling}, Pseudo-label~\cite{pseudolabel} and FixMatch~\cite{fixmatch}.
	More detailed descriptions are given in the supplementary.
	
	\subsection{Metrics}
	\label{subsec:metric}
	
	The evaluation protocols vary slightly depending on the used output and dataset.
	The used data splits training, validation and unlabeled are defined above in Section \ref{subsec:datasets}.
	
	On STL-10, we calculate accuracy of the validation data.
	Accuracy is the portion of true positive and true negatives from the complete dataset.
	\begin{equation}
	    Accuracy = \frac{TP + TN}{TP + TN + FP + FN}
	\end{equation}
	
	TP, TN, FP and FN are the true positive, true negative, false positive and false negative respectively.
	We calculate these values per class and then sum the up before calculating the accuracy (micro averaging).
	For the overclustering head, we need to find a mapping between the output clusters and the given classes.
	We calculate this mapping based on the majority class in each cluster on the training data as in~\cite{iic}.
	
	On the \fuzzy plankton and synthetic datasets, we evaluate the macro-f1 score on the unlabeled data.
	We calculate the macro F1-score i.e. the average of the F1-scores per class  due to the skewed class distribution.
	\begin{equation}
	    F1-Score = \frac{2TP}{2TP + FP + FN}
	\end{equation}
	
	Mind that a micro averaged F1-Score would be in our case the same as the above defined accuracy.
	We use the unlabeled data as evaluation dataset because the \fuzzy images, in which we are interested, are only included in the unlabeled data split by definition.
	The mapping for the overclustering head is calculated based on the unlabeled data split because we expect human experts to be involved in this process for the identification of substructures.
	The best unlabeled results of the \fuzzy Plankton and Synthetic dataset are reported based on the validation metrics.
	
	If not stated otherwise, we report the maximum score for the overclustering and the normal head and the average and standard deviation over 3 independent repetitions.

	\subsection{Results}
	\label{subsec:experiments}

	\subsubsection{State-of-the-Art Comparison}	
	
	We compare the state-of-the-art methods on \certain and \fuzzy data in Table \ref{tbl:comparisonPlankton}.

		\begin{table}
		\centering

			\begin{tabular}{l  c c c  }
				\toprule
					& & \multicolumn{2}{c}{Type of Data}  \\
				\cmidrule(r){3-4} 
				Method & Network & Certain & Fuzzy \\
				
				\midrule
				
				SCAN $^\star$ \cite{scan} & ResNet18 & 76.80  $\pm$ 1.10 & 37.64 $\pm$ 3.56 \\	
				IIC \cite{iic} & ResNet34 & 85.76 $\pm$ 1.36 & 65.47 $\pm$ 1.86    \\
				IIC $^\dagger$ \cite{iic} & ResNet34 & 88.8 & 66.81 $\pm$ 1.85    \\				
				Mean-Teacher \cite{mean-teacher}  & Wide ResNet28  & 78.577  $\pm$ 2.39 $^\ddagger$ & 72.85  $\pm$ 0.46   \\
				Pi \cite{temporal-ensembling} & Wide ResNet28  & 73.77  $\pm$  0.82 $^\ddagger$& 74.34  $\pm$  0.58  \\
				Pseudo-label \cite{pseudolabel} & Wide ResNet28  &  72.01 $\pm$ 0.83  $^\ddagger$& 75.04  $\pm$ 0.52 \\
				FixMatch \cite{fixmatch} & Wide ResNet28  &  \textbf{94.83 $\pm$ 0.63 $^\ddagger$ }& 76.28  $\pm$ 0.27  \\	
				FOC-Light (Ours) & ResNet50 &  -- & 72.79   $\pm$ 2.99\\
				FOC (Ours) &  ResNet50 &  86.12 $\pm$ 1.22 & \textbf{76.79  $\pm$  1.18}  \\

				\bottomrule
			\end{tabular}	
		
		\caption{Comparison of state-of-the-art on certain and fuzzy data---We use STL-10 as a \certain dataset and the Plankton data as a \fuzzy dataset.
    	We report the Accuracy for STL-10 and the F1-Score for the Plankton data due to class imbalance.
    	It is important to notice that STL-10 is a curated dataset while the Plankton dataset still contains the \fuzzy  images.
    	For more details about the metrics see Section \ref{subsec:metric}.
    		The results of previous methods are reported in the original paper or the original authors code was used to replicate the results.
		 	The best results are marked bold.		 	 
			Legend:
			$^\dagger$ A MLP used for fine-tuning.
			$^\ddagger$ Used only 1000 labels instead of 5000.
			$^\star$ Unsupervised method.
		}
		\label{tbl:comparisonPlankton}
		
	\end{table}

	We see that FOC reaches a performance of about 86\%  on \certain data but is not able to reach the performance of FixMatch. 
	FixMatch outperforms FOC by a clear margin of nearly 8\% while using a fifth of the labels.
	This performance is expected as FOC does not focus like the others on classifying certain but \fuzzy data.
	If we look at the less curated \fuzzy Plankton dataset, we see that FOC  outperforms all all methods by a small margin.
	All previous methods focus on certain and curated data and we see this leads to a huge performance degeneration if they are applied to \fuzzy data.
	FixMatch reaches in both datasets the best performance except for our method FOC.
	We conclude that the overclustering from FOC is the key for handling \fuzzy data because it allows more flexibility during training.
	Previous semi-supervised methods did not consider the issue of inter- and intraobserver variability and thus are worse than FOC in classifying \fuzzy data. 
	
	If we use FOC-Light without the loss and stabilization of~\cite{iic} the F1-Score drops slightly to 75\% but the used GPU hours can be decreased from 58 to 4 h.
	We conclude that the overclustering head is more suitable for handling \fuzzy real-world data as we assumed at the beginning.
	Moreover, we see that the combination of cross-entropy and our novel loss \ceinv~can also successfully train an overclustering head.
	
	\subsubsection{Consistency}
	\label{subsec:consistency}
	
	Up to this point, we analyzed classification metrics based on the 10 ground-truth classes but the quality of substructures was not evaluated.
	We can judge the consistency of each image within its cluster with the help of experts as a quality measure.
	An image is consistent if an expert views it as visually similar to the majority of the cluster.
	The consistency is calculated by dividing the number of consistent images by all images. 
	The consistency over all classes or per class for FOC and FixMatch is given in Table \ref{tbl:consistency} and raw numbers are provided in the supplementary.
	We provide a comparison based on all data and without the no-fit class because this class contains a mixture of different plankton entities.
	Visual similarity is therefore difficult to judge because it can only be defined by not being similar the other nine classes.
	Based on the F1-Score, FixMatch and FOC perform similarly but if we look at the consistency we see that FOC is more than 5\% more consistent than FixMatch.
	If we exclude the class no-fit from the analysis, FOC reaches a consistency of around 86\% in comparison to  77\% from FixMatch.
	For both sets, our method FOC reaches a higher average consistency per cluster and lower standard deviation.
	This means the clusters produced by FOC are more relevant in practice because there are fewer low-quality clusters which can not be used.
	Overall, this higher consistency can lead to faster and more reliable annotations.

\begin{table}

		\centering
		
		\begin{tabular}{l c c c c }
			\toprule
	
			& \multicolumn{2}{c}{all data}  & 	\multicolumn{2}{c}{ignore class no-fit} \\ 
			
			\cmidrule(r){2-3} \cmidrule(r){4-5}
			Method  & overall & per cluster & overall & per cluster \\
			\midrule

			FixMatch \cite{fixmatch} &  82.56 & 78.78 $\pm$ 28.22 & 77.11 & 69.61 $\pm$ 29.41 \\
			FOC (Ours) &\textbf{87.80} & \textbf{79.66 $\pm$ 18.88} & \textbf{86.31} & \textbf{86.41 $\pm$ 13.68}\\
			
			\bottomrule
		\end{tabular}
		
		\caption{Consistency comparison on plankton dataset --
		    The consistency is rated by experts over the complete data and a subset without the class no-fit.
		    The score is given overall as as average per cluster with standard deviation and is described in \autoref{subsec:consistency}.
		    }
		 \label{tbl:consistency} 
		   
\end{table}

	\subsubsection{Qualitative Results}
	
\begin{figure}
\begin{tabular}{cccccccccc}

\includegraphics[width=0.077\textwidth]{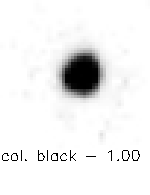}&\includegraphics[width=0.077\textwidth]{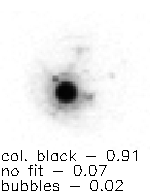}&\includegraphics[width=0.077\textwidth]{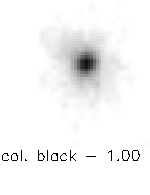}&\includegraphics[width=0.077\textwidth]{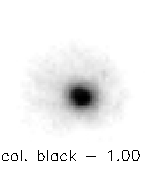}&\includegraphics[width=0.077\textwidth]{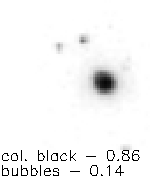}&\includegraphics[width=0.077\textwidth]{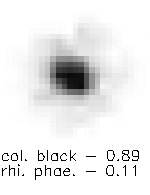}&\includegraphics[width=0.077\textwidth]{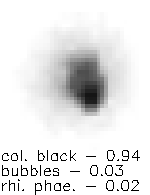}&\includegraphics[width=0.077\textwidth]{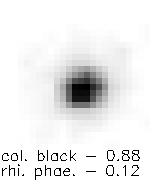}&\includegraphics[width=0.077\textwidth]{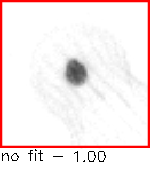}&\includegraphics[width=0.077\textwidth]{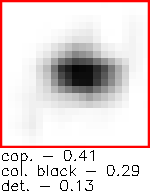}\\
\includegraphics[width=0.077\textwidth]{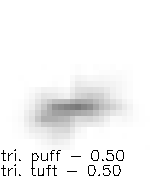}&\includegraphics[width=0.077\textwidth]{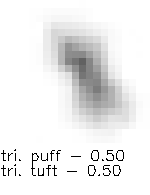}&\includegraphics[width=0.077\textwidth]{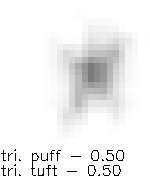}&\includegraphics[width=0.077\textwidth]{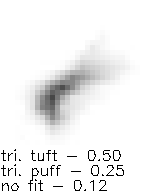}&\includegraphics[width=0.077\textwidth]{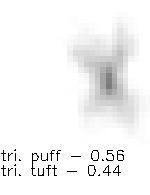}&\includegraphics[width=0.077\textwidth]{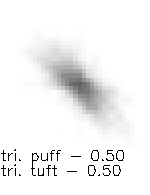}&\includegraphics[width=0.077\textwidth]{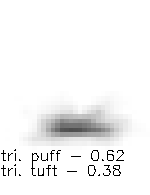}&\includegraphics[width=0.077\textwidth]{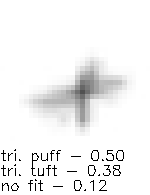}&\includegraphics[width=0.077\textwidth]{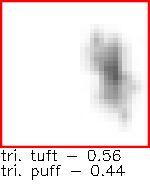}&\includegraphics[width=0.077\textwidth]{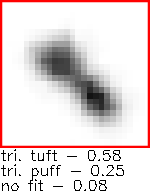}

\end{tabular}
\caption{Qualitative results for unlabeled data---The results in each row are from the same predicted cluster.
		The three most important \fuzzy labels based on the citizen scientists' annotations are given below the image. 
	The last two items with the red box in each row show examples not matching the majority of the cluster.}\label{fig:qualitative}
\end{figure}

	We illustrate some qualitative results of FOC in Figure \ref{fig:qualitative}.
	All images in a cluster are visually similar, even the probably wrongly assigned images (red box).
	For the images in the first row, the annotators are certain that the images belong to the same class.
	In the second row, annotators show a high uncertainty of assignment between the two variants of the same biological object.
	This illustrates the benefit of overclustering since visual similar items are in the same cluster even for uncertain annotations.
	In a consensus process for the second row, experts could decide if the cluster should be the puff, tuft or a new borderline class.
	Moreover, this clustering could be beneficial for monitoring the current imaging process.
	We provide more randomly selected results in the supplementary.
	
	\subsection{Ablation Studies}	

\begin{table}
		\centering

			\begin{tabular}{l  c c c c}
				\toprule
				Method	& Ideal &  Real & Fuzzy\\
				
				\midrule

Mean-Teacher \cite{mean-teacher}  & 97.11 $\pm$ 0.78  & 73.23 $\pm$ 2.49 & 66.57 $\pm$ 16.27   \\
				Pi \cite{temporal-ensembling} &\textbf{ 98.44 $\pm$ 0.28 }& 72.74 $\pm$ 2.43 & 77.69 $\pm$ 5.02  \\
				Pseudo-label \cite{pseudolabel} & \textbf{98.17 $\pm$ 0.30} & 75.70 $\pm$ 1.98 & 89.48 $\pm$ 1.94 \\
				FixMatch \cite{fixmatch} &\textbf{ 98.32 $\pm$ 0.01} & 71.81  $\pm$ 1.06 & \textbf{93.82  $\pm$ 1.83}\\		
				FOC-Light (Ours) &  \textbf{97.46 $\pm$ 4.39 }& 78.77 $\pm$ 7.83 & \textbf{94.29 $\pm$ 0.87} \\
				FOC (Ours) &\textbf{ 97.72 $\pm$ 4.52} & \textbf{83.86 $\pm$ 4.21} & \textbf{94.15 $\pm$ 0.29 }  \\

				\bottomrule
			\end{tabular}	
		
		\caption{Comparison to state-of-the-art on SYN-CE datasets -- 
			Each column represent a subset of the dataset SYN-CE.			
			The results are F1-Scores which were calculated on the unlabeled data which include the fuzzy labels.
			All results within a one percent margin of the best result are marked bold.
		}
		\label{tbl:comparisonSYN}
		
	\end{table}

\begin{table}
	\centering
	
	\begin{minipage}{\linewidth}
	\centering
		\begin{tabular}{l c c c c c c}
			\toprule
			& & & & & \multicolumn{2}{c}{Accuracy}     \\ 
			\cmidrule(r){6-7}
			Method & Warm & MI & \ceinv &  Weight &    Overcluster & Normal \\
			\midrule
		FOC  &  & X& & -- &   70.92 $\pm$ 2.42 & 76.39 $\pm$ 0.05 \\    
			IIC * \cite{iic} &  X  & & & -- &   & 85.76 \\				
			FOC  &  X & X &  & --  & 73.88 $\pm$ 0.21 & 82.01 $\pm$ 5.31\\
			FOC &  X & X& X & -- &  82.59 $\pm$ 0.06 & 86.49 $\pm$ 0.01 \\      	
			FOC  &  X & X & X & C &    84.36  $\pm$  0.64 & 78.59  $\pm$ 7.40 \\
			FOC  & X & X&X & I  & 83.57 $\pm$ 0.10 & 85.21 $\pm$ 0.03 \\
			
			\bottomrule
		\end{tabular}
	\subcaption{STL-10}
	
	\end{minipage}%
	\hfill
	\begin{minipage}{\linewidth}
	\centering
			\begin{tabular}{l c c c c c c}
				\toprule
				& & & & &   \multicolumn{2}{c}{F1-Score}     \\ 				
			  \cmidrule(r){6-7} 
				Method & Warm & MI & \ceinv &  Weight &    Overcluster & Normal  \\ 
				\midrule
				
				IIC \cite{iic}  & X &  &  & -- &  -- & 66.63 \\
				IIC $^\dagger$ \cite{iic} & X &  &  & -- &   -- & 69.92  \\		
				
				\midrule
				
				FOC & & &  & C  & 31.45  $\pm$  6.02 & 39.35  $\pm$  1.30 \\
				FOC & & X & & C &  29.82 $\pm$ 2.98 &  60.65 $\pm$ 0.02\\
				FOC & & X & X & C  & 70.11 $\pm$ 1.99 &  64.10 $\pm$ 0.13 \\
				
				\midrule
				FOC & X & &  & C & 23.95 $\pm$ 2.63  & 58.71 $\pm$ 2.07 \\
				FOC& X & X& & C &  69.36 $\pm$ 0.05 & 56.59 $\pm$ 0.04 \\
				FOC & X & X & X & C  & 70.68 $\pm$ 0.10 &  58.09 $\pm$ 0.03\\
				
				\midrule
				
				FOC & & & & I &  29.88  $\pm$  2.75 & 54.92  $\pm$  0.03 \\
				FOC-Light & & & X & I & 74.93 $\pm$ 0.22 & 73.64 $\pm$ 0.06\\			  
				FOC & & X & & I &  72.70 $\pm$ 0.36 & 64.78 $\pm$ 0.04 \\
				FOC &  & X& X & I &  73.93 $\pm$ 0.29 & 64.84 $\pm$ 0.03 \\
				
				\midrule
				FOC & X & &  & I &  73.93 $\pm$ 0.29 & 64.84 $\pm$ 0.03 \\
				FOC & X & X & & I &  69.64 $\pm$ 1.04 &  66.56 $\pm$ 0.08 \\
				FOC & X & X & X & I &  74.01 $\pm$ 3.17 &  65.17 $\pm$ 0.18\\
				
				\bottomrule
			\end{tabular}
		
		\subcaption{plankton dataset}

	\end{minipage}%
	
	 \caption{Ablation study --
		The second to fourth column indicates if a warm-up, the MI loss or our \ceinv loss  were used respectively.
		The fifth column indicates if CIFAR-20 (C), ImageNet (I) or no  (-) weights were used.
		Sobel filtered images are used as input for no weights.
		The Top1 and Top3 results are marked bold respectively.
		 * Original authors code 
			$^\dagger$ A MLP used for fine-tuning.
		}	
			\label{tbl:Ablation}	
\end{table}

	\subsubsection{SYN-CE}
	\label{subsec:syn}
	
	We compare our framework with some previous methods on the three subsets of SYN-CE in Table \ref{tbl:comparisonSYN}.
	All semi-supervised method reach almost a F1-Score of 100\% on the unlabeled \fuzzy data for the subset \emph{Ideal}.
	In real-world data, it is unlikely that we have the real \fuzzy ground-truth labels. 
	It is more likely that we have uncertain/wrong labels for training and validation or no labels at all for \fuzzy data like in the subsets \emph{Real} or \emph{Fuzzy}.
	In both cases, we see that our method reaches a superior performance with up to 10\% higher F1-Score.
	While FOC-Light is only slightly better in comparison to the other semi-supervised methods on the \emph{Real} subset it is comparable to the complete framework on the \emph{Fuzzy} dataset. 
	This is one indication that \ceinv~is one of the key components for successfully training the overclustering heads.
	We see the F1-Score on the \emph{Fuzzy} subset is around 10\% higher than on the \emph{Real} subset.
	We conclude that FOC can also generalize to other datasets.
	We conclude that these results supports our idea of separating \certain and \fuzzy data during training because we do not need to potentially falsely approximate the real \fuzzy ground-truth label like in the \emph{Real} subset.

	\subsubsection{Loss \& Network}
	
	In Table \ref{tbl:Ablation} multiple ablations for STL-10 and the plankton dataset are given.
	The scores are averaged across the different output heads of our framework.
	Based on these tables, we illustrate the impact of the warm-up, the initialization and the usage of the MI and \ceinv~loss for our framework.	
	The normal accuracy can be improved by about 10\% when using the unsupervised warm-up on the STL-10 dataset.
	On the plankton dataset, the impact is less but tends to give better results of some percent.
	Warm-up in combination with the MI loss leads to a performance which is not more than 10\% worse than the full setup for all ablations except for one.
	For this exception, \ceinv~is needed to stabilize the overclustering performance due to the poor initialization with CIFAR-20 weights.
	We attribute this worse performance to the initialization and not the different backbone because on STL-10 the CIFAR-20 initializations of the ResNet34 backbone outperform the ImageNet weights of the ResNet50v2 backbone.
	We believe the positive effects of ImageNet weights for its subset STL-10 and the better network are negated by the different loss.

	IIC is similar to FOC with warm-up and no additional losses but we train also train an overclustering head for handling \fuzzy data.
	Taking this into consideration, we achieve an 8 to 11\% better F1-Score than IIC.
	A special case is FOC-light which does only use the \ceinv~loss and therefore no stabilization method proposed in~\cite{iic}.
	This decreases gpu memory usage and runtime and results in a total decrease of the GPU hours from 58 to 4~h.
	Overall, our novel loss \ceinv~improves the overclustering performance regardless of the dataset and the weight initialization by 10\% on STL-10 and up to 7\% on the plankton dataset.
	We see that \ceinv~is a key component for training an overclustering head and can even be trained without the stabilization of the warm-up and the MI loss.

\

	\section{Conclusions}
	
	In this paper, we take the first steps to address real-world underwater issues with semi-supervised learning.
	Our presented novel framework FOC can handle \fuzzy labels via overclustering.
	We showed that overclustering can achieve better results than previous state-of-the-art semi-supervised methods on \fuzzy plankton data.
	The additional overclustering output is a key difference to previous work to achieve this superior performance.
	While on \certain data FOC is not state-of-the-art by a clear margin of over 10\%, it slighlty outperforms all other methods on the \fuzzy plankton data.
	These beneficial effects have to be verified on other \fuzzy datasets and with more semi-supervised algorithms in the future.
	Due to better performance of FOC on \fuzzy data, we expect a similar outcome.
	We  illustrated the visual similarity on qualitative results from these predictions and results in 5 to 10\% more consistent predictions.
	We showed that \ceinv~is the key component for training the overclustering head.

\bibliographystyle{IEEEtranN}
\bibliography{lib}

\end{document}